\documentclass[10pt,twocolumn,letterpaper]{article}

\usepackage{wacv}
\usepackage{times}
\usepackage{epsfig}
\usepackage{graphicx}
\usepackage{amsmath}
\usepackage{amssymb}

\usepackage{multirow}
\usepackage{url}
\usepackage{subfigure}
\usepackage{xcolor}
\usepackage{float}
\usepackage{makecell}
\usepackage{color, colortbl}
\usepackage{booktabs}
\usepackage[bookmarks=false]{hyperref}
\usepackage{ulem}

\definecolor{Gray}{gray}{0.9}
\definecolor{LightCyan}{rgb}{0.88,1,1}

\wacvfinalcopy 

\def\wacvPaperID{137} 

\ifwacvfinal\fi
\begin{document}

\title{A Hierarchical Grocery Store Image Dataset with Visual and Semantic Labels}

 \author{Marcus Klasson$^1$\thanks{This research is funded by the Promobilia foundation and the Swedish e-Science Research Centre.}~~~~~~
 Cheng Zhang$^2$ ~~~~~~ Hedvig Kjellstr\"{o}m$^1$\vspace{2mm}
 \\
 $^1$ KTH Royal Institute of Technology, Stockholm, Sweden, 
 {\normalsize {\tt \{mklas,hedvig\}@kth.se}} \\
 $^2$ Microsoft Research, Cambridge, United Kingdom, {\normalsize {\tt cheng.zhang@microsoft.com}}
 }

\maketitle
\ifwacvfinal\thispagestyle{empty}\fi

\begin{abstract}
Image classification models built into visual support systems and other assistive devices need to provide accurate predictions about their environment. We focus on an application of assistive technology for people with visual impairments, for daily activities such as shopping or cooking. In this paper, we provide a new benchmark dataset for a challenging task in this application -- classification of fruits, vegetables, and refrigerated products, e.g. milk packages and juice cartons, in grocery stores. To enable the learning process to utilize multiple sources of structured information, this dataset not only contains a large volume of natural images but also includes the corresponding information of the product from an online shopping website. Such information encompasses the hierarchical structure of the object classes, as well as an iconic image of each type of object. This dataset can be used to train and evaluate image classification models for helping visually impaired people in natural environments. Additionally, we provide benchmark results evaluated on pretrained convolutional neural networks often used for image understanding purposes, and also a multi-view variational autoencoder, which is capable of utilizing the rich product information in the dataset.
\end{abstract}

\section{Introduction}
\label{sec:introduction}

In this paper, we focus on the application of image recognition models implemented into assistive technologies for people with visual impairments. Such technologies already exist in the form of mobile applications, e.g.~Microsoft's Seeing AI \cite{seeingAImicrosoft} and Aipoly Vision \cite{aipolyvision}, and as wearable artificial vision devices, e.g.~Orcam MyEye \cite{orcam} and the Sound of Vision system introduced in \cite{caraiman2017soundofvision}. These products have the ability to support people with visual impairments in many different situations, such as reading text documents, describing the user's environment and recognizing people the user may know. 

\begin{figure}[t]
	\centering
    \includegraphics[width=\columnwidth]{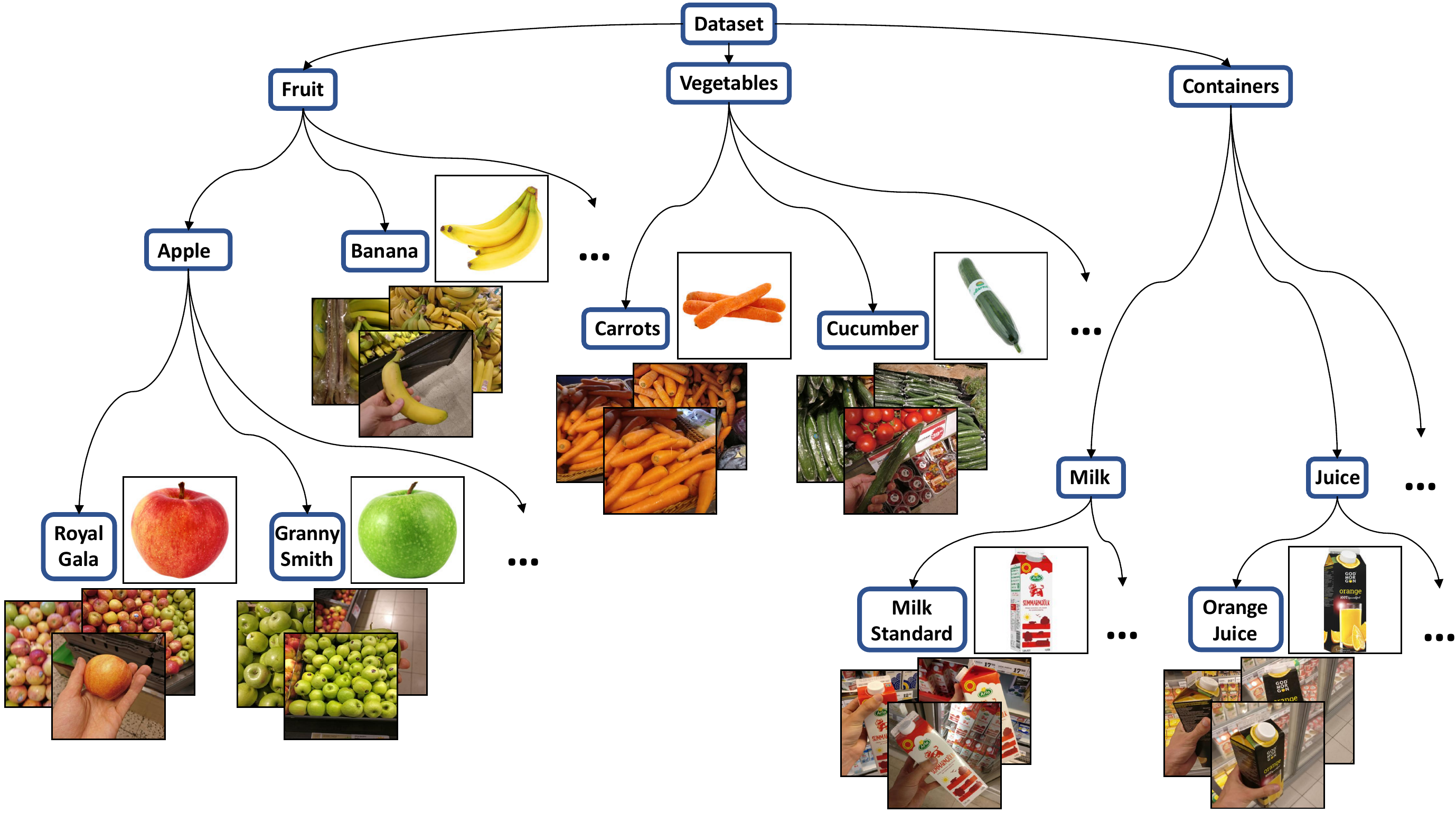}
    \vspace{-2mm}
    \caption{The primary contribution of this paper is a dataset of grocery items, for the purpose of training a visual recognition system to aid visually impaired people. The dataset is organized according to a hierarchical class structure, as illustrated above. A novel aspect of the dataset is that each class, apart from the semantic label, also has a visual label in the form of an iconic image.}
    \label{fig:examples} 
\end{figure}

\begin{figure*}[t] 
\centering
\begin{minipage}[t]{0.47\textwidth}
\centering
\subfigure[Royal Gala]{\label{subfig:real-image-a}\includegraphics[width=0.30\columnwidth]{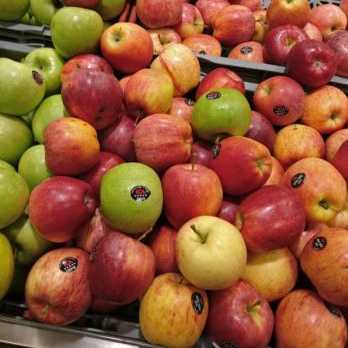}}~
\subfigure[Golden Delicious]{\label{subfig:real-image-c}\includegraphics[width=0.30\columnwidth]{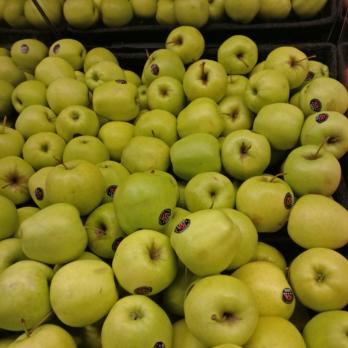}}~
\subfigure[Orange]{\label{subfig:real-image-f}\includegraphics[width=0.30\columnwidth]{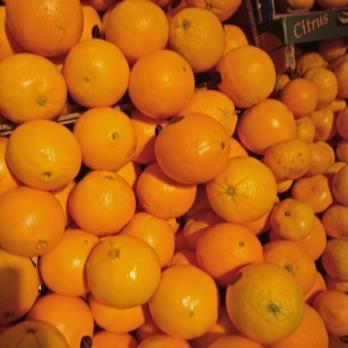}}~ \\ 
\subfigure[Aubergine]{\label{subfig:real-image-h}\includegraphics[width=0.30\columnwidth]{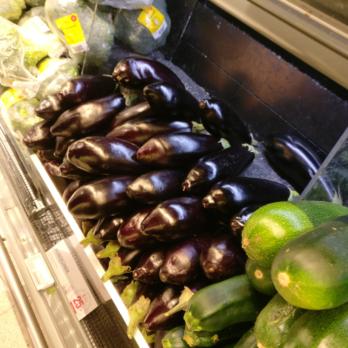}}~
\subfigure[Onion]{\label{subfig:real-image-j}\includegraphics[width=0.30\columnwidth]{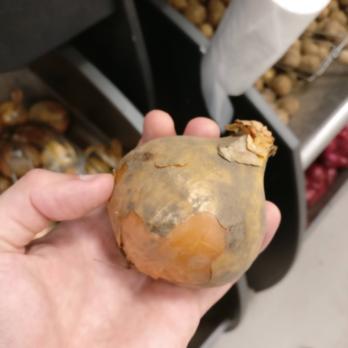}}~
\subfigure[Zucchini]{\label{subfig:real-image-l}\includegraphics[width=0.30\columnwidth]{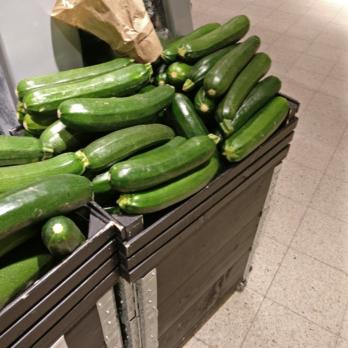}}~ \\ 
\subfigure[Apple Juice]{\label{subfig:real-image-n}\includegraphics[width=0.30\columnwidth]{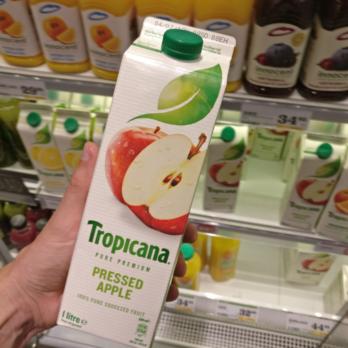}}~
\subfigure[Milk Medium Fat]{\label{subfig:real-image-q}\includegraphics[width=0.30\columnwidth]{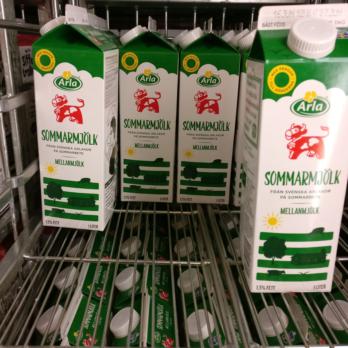}}~
\subfigure[Yogurt Natural]{\label{subfig:real-image-r}\includegraphics[width=0.30\columnwidth]{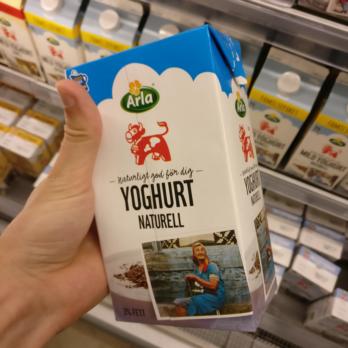}}~
   \caption{Examples of natural images in our dataset, where each image have been taken inside a grocery store. Image examples of fruits, vegetables and refrigerated products are presented in each row respectively.
   }
\label{fig:dataset-figure}
\end{minipage}
\hspace{10pt}
\begin{minipage}[t]{0.47\textwidth}
\vspace{0pt}
\centering
\subfigure[Royal Gala]{\label{subfig:clean-img-a}\includegraphics[width=0.30\columnwidth]{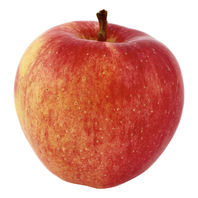}}~
\subfigure[Golden Delicious]{\label{subfig:clean-img-c}\includegraphics[width=0.30\columnwidth]{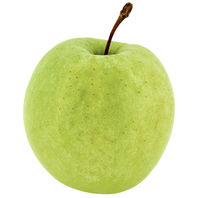}}~
\subfigure[Orange]{\label{subfig:clean-img-f}\includegraphics[width=0.30\columnwidth]{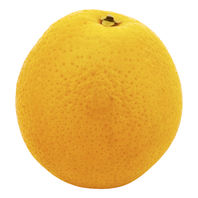}}~ \\ 
\subfigure[Aubergine]{\label{subfig:clean-img-h}\includegraphics[width=0.30\columnwidth]{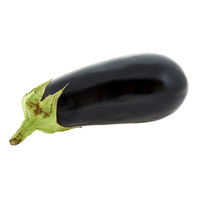}}~
\subfigure[Onion]{\label{subfig:clean-img-j}\includegraphics[width=0.30\columnwidth]{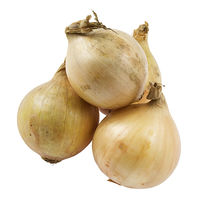}}~
\subfigure[Zucchini]{\label{subfig:clean-img-l}\includegraphics[width=0.30\columnwidth]{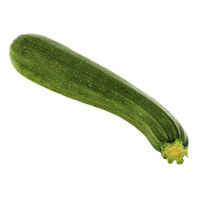}}~ \\ 
\subfigure[Apple Juice]{\label{subfig:clean-img-n}\includegraphics[width=0.30\columnwidth]{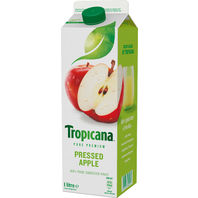}}~
\subfigure[Milk Medium Fat]{\label{subfig:clean-img-q}\includegraphics[width=0.30\columnwidth]{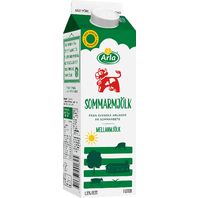}}~
\subfigure[Yogurt Natural]{\label{subfig:clean-img-r}\includegraphics[width=0.30\columnwidth]{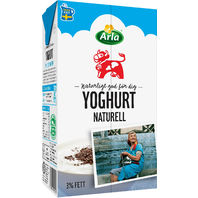}}~
   \caption{Examples of iconic images downloaded from a grocery shopping website, which corresponds to the target items in the images in Figure \ref{fig:dataset-figure}.}
\label{fig:clean-image-figure}
\end{minipage}
\end{figure*}

We here address a complementary scenario not handled by current systems on the market: visual support when shopping for grocery items considering a large range of eatable objects, including fruits, vegetables, milk, and juices. 
In the case of fruits and vegetables, these are usually stacked in large bins in grocery stores as shown in Figure \ref{fig:dataset-figure}(a-f). A common problem in grocery stores is that similar items are often stacked next to each other; therefore, items are often misplaced into neighboring bins. Figure \ref{subfig:real-image-a} shows a mix of red and green apples, where it might be difficult for the system to determine which kind of apple is the actual target.
Humans can distinguish between groceries without vision to some degree, e.g.~by touching and smelling them, but it requires prior knowledge about texture and fragrance of food items.

Moreover, in addition to raw grocery items, there are also items that can only be differentiated with the help of visual information, e.g. milk, juice, and yogurt cartons, see Figure \ref{fig:dataset-figure}(g-i). Such items usually have barcodes, that are readable using the existing assistive devices described above. However, the barcodes are not easily located by visually impaired persons. Thus, an assistive vision device that fully relies on natural image understanding would be of significant added value for a visually impaired person shopping in a grocery store.

Image recognition models used for this task typically require training images collected in similar environments. However, current benchmark datasets, such as ImageNet \cite{deng2009imagenet} and CIFAR-100 \cite{Krizhevsky2009cifar100}, do contain images of fruits and vegetables, but are not suitable for this type of assistive application, 
since the target objects are commonly not presented in this type of natural environments, with occlusion and cluttered backgrounds. To address this issue, we present a novel dataset containing natural images of various raw grocery items and refrigerated products, e.g. milk, juice, and yogurt, taken in grocery stores. As part of our dataset, we collect images taken with single and multiple target objects, from various perspectives, and with noisy backgrounds.

In computer vision, previous studies have shown that model performance can be improved by extending the model to utilize other data sources, e.g. text, audio, in various machine learning tasks \cite{frome2013DeVISE,Gebru2017FineGrainedCD,karpathy2015deepvisualsemantic,ngiam2011multimodal}. Descriptions of images are rather common to computer vision datasets, e.g. Flickr30k \cite{plummer2015flickr30k}, whereas the datasets in \cite{Gebru2017FineGrainedCD,Lin2014MicrosoftCoco} includes both descriptions and a reference image with clean background to some objects. Therefore, in addition to the natural images, we have collected iconic images with a single object centered in the image (see Figure \ref{fig:clean-image-figure}) and a corresponding product description to each grocery item. In this work, we also demonstrate how we can benefit from using additional information about the natural images by applying the multi-view generative model.

To summarize, the contribution of this paper is a dataset of natural images of raw and refrigerated grocery items, which could be used for evaluating and training image recognition systems to assist visually impaired people in a grocery store. 
The dataset labels have a hierarchical structure with both coarse- and fine-grained classes (see Figure \ref{fig:examples}). Moreover, 
each class also has an iconic image and a product description, which makes the dataset applicable to multimodal learning models. The dataset is described in Section \ref{sec:our-dataset}. 

We provide multiple benchmark results using various deep neural networks, such as Alexnet \cite{krizhevsky2012imagenet}, VGG \cite{simonyan2014verydeep}, DenseNet \cite{huang2017densely}, as well as deep generative models, such as VAE \cite{kingma2014autoencoding}. 
Furthermore, we adapt a multi-view VAE model to make use of the iconic images for each class (Section \ref{sec:classification-methods}), and show that it improves the classification accuracy given the same model setting (Section \ref{sec:experimental-results}). Last, we discuss possible future directions for fully using the additional information provided with the dataset and adopt more advanced machine learning methods, such as visual-semantic embeddings, to learn efficient representations of the images. 

\section{Related Work}\label{sec:related-work}

Many popular image datasets have been collected by downloading images from the web \cite{deng2009imagenet,Everingham2010pascal,Gebru2017FineGrainedCD, griffin2007caltech256,Krizhevsky2009cifar100,Lin2014MicrosoftCoco, song2016deep, welinder2010birds,xiao2010sundatabase}. 
If the dataset contains a large amount of images, it is convenient to make use of crowdsourcing to get annotations for recognition tasks \cite{deng2009imagenet,Krizhevsky2009cifar100,liu2015faceattributes}. For some datasets, the crowdsourcers are also asked to put bounding boxes around the object to be labeled for object detection tasks \cite{Everingham2010pascal,Gebru2017FineGrainedCD,welinder2010birds}. In \cite{griffin2007caltech256} and \cite{Krizhevsky2009cifar100}, the target objects are usually centered and takes up most content of the image itself. Another significant characteristic is that web images usually are biased in the sense that they have been taken with the object focus in mind; they have good lighting settings and are typically clean from occlusions, since the collectors have used general search words for the object classes, e.g. \textit{car}, \textit{horse}, or \textit{apple}.

Some datasets include additional information about the images beyond the single class label, e.g. text descriptions of what is present in the image and bounding boxes around objects. These datasets can be used in several different computer vision tasks, such as image classification, object detection, and image segmentation. Structured labeling is another important property of a dataset, which provides flexibility when classifying images. In  \cite{Gebru2017FineGrainedCD,Lin2014MicrosoftCoco}, all of these features exist and moreover they include reference images to each object class, which in \cite{Lin2014MicrosoftCoco} is used for labeling multiple  categories present in images, while in \cite{Gebru2017FineGrainedCD} these images are used for fine-grained recognition. 
Our dataset includes a reference image, i.e. the iconic image, and a product description for every class, and we have also labeled the grocery items in a structured manner.

Other image datasets of fruits and vegetables for classification purposes are the FIDS30 database \cite{marko2013fids30} and the dataset in \cite{muresan2017fruit}. The images in FIDS30 were downloaded from the web and contain background noise as well as single or multiple instances of the object. In \cite{muresan2017fruit}, all pixels belonging to the object are extracted from the original image, such that all images have white backgrounds with the same brightness condition. There also exist datasets for detecting fruits in orchards for robotic harvesting purposes, which are very challenging since the images contain plenty of background and various lighting conditions, and the targeted fruits are often occluded or of the same color as the background \cite{bargoti2017deepfruitdetection,sa2016deepfruits}.

Another dataset that is highly relevant to our application need is presented in  \cite{waltner2015mango}. They collected a dataset for training and evaluating the image classifier by extracting images from video recordings of 23 main classes, which are subdivided into 98 classes, of raw grocery items (fruits and vegetables) in different grocery stores. Using this dataset, a mobile application was developed to recognize food products in grocery store environments, which provides the user with details and health recommendations about the item along with other proposals of similar food items. For each class, there exists a product description with nutrition values to assist the user in shopping scenarios. The main difference between this work and our dataset is firstly the clean iconic images (visual labels) for each class in our dataset, and secondly that we have also collected images of refrigerated items, such as dairy and juice containers, where visual information is required to distinguish between the products.   

\section{Our Dataset}\label{sec:our-dataset}

\begin{figure}[t]
\centering
\includegraphics[width=\columnwidth,height=0.20\paperheight]{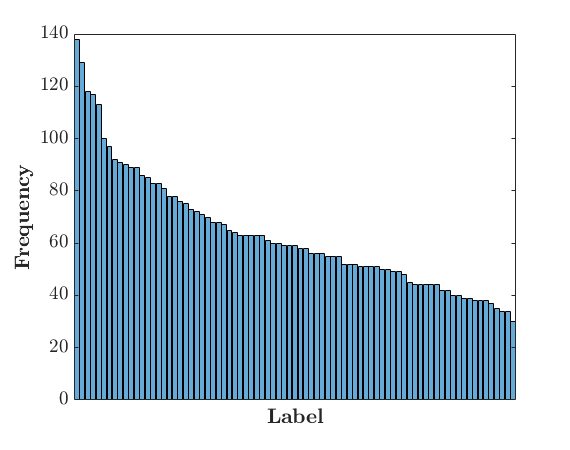}
\caption{Histogram over the number of images in each class in the dataset.}
\label{fig:hist}
\end{figure}

We have collected images from fruit and vegetable sections and refrigerated sections with dairy and juice products in 18 different grocery stores. The dataset consists of 5125 images from 81 fine-grained classes, where the number of images in each class range from 30 to 138. Figure \ref{fig:hist} displays a histogram over the number of images per class. As illustrated in Figure \ref{fig:examples}, the class structure is hierarchical, and there are 46 coarse-grained classes. Figure \ref{fig:dataset-figure} shows examples of the collected natural images. For each fine-grained class, we have downloaded an iconic image of the item and also a product description including origin country, an appreciated weight and nutrient values of the item from a grocery store website. Some examples of downloaded iconic images can be seen in Figure \ref{fig:clean-image-figure}. 

Our aim has been to collect the natural images under the same condition as they would be as part of an assistive application on a mobile phone. All images have been taken with a 16-megapixel Android smartphone camera from different distances and angles. Occasionally, the images include other items in the background or even items that have been misplaced in the wrong shelf along with the targeted item. It is important that image classifiers that are used for assisting devices are capable of performing well with such noise since these are typical settings in a grocery store environments. The lighting conditions in the images can also vary depending on where the items are located in the store. 
Sometimes the images are taken while the photographer is holding the item in the hand. This is often the case for refrigerated products since these containers are usually stacked compactly in the refrigerators. For these images, we have consciously varied the position of the object, such that the item is not always centered in the image or present in its entirety. 

We also split the data into a training set and test set based on the application need. Since the images have been taken in several different stores at specific days and time stamps, 
parts of the data will have similar lighting conditions and backgrounds for each photo occasion. To remove any such biasing correlations, all images of a certain class taken at a certain store are assigned to either the test set or training set. Moreover, we balance the class sizes to as large extent as possible in both the training and test set. After the partitioning, the training and test set contains 2640 and 2485 images respectively. Predefining a training and test set also makes it easier for other users to compare their results to the evaluations in this paper.

The task is to classify natural images using mobile devices to aid visually impaired people. The additional information such as the hierarchical structure of the class labels, iconic images, and product descriptions can be used to improve the performance of the computer vision system. Every class label is associated with a product description. Thus, the product description itself can be part of the output for visually impaired persons as they may not be able to read what is printed on a carton box or a label tag on a fruit bin in the store.

The dataset is intended for research purposes and we are open to contributions with more images and new suitable classes. Our dataset is available at \url{https://github.com/marcusklasson/GroceryStoreDataset}. Detailed instructions on how to contribute to the dataset can be found on our dataset webpage.

\section{Classification Methods}\label{sec:classification-methods}

We here describe the classification methods and approaches that we have used to provide benchmark results to the dataset. We apply both deterministic deep neural networks as well as a deep generative model used for representation learning to the natural images that we have collected. Furthermore, we utilize the additional information -- iconic images -- from our dataset with a multi-view deep generative model. This model can utilize different data sources and obtain superior representation quality as well as high interpretability. For a fair evaluation, we use a linear classifier with the learned representation from the different methods.

\paragraph*{Deep Neural Networks.} 
CNNs have been the state-of-the-art models in image classification ever since AlexNet \cite{krizhevsky2012imagenet} achieved the best classification accuracy in ILSVRC in 2012.
However, in general, computer vision models require lots of labeled data to achieve satisfactory performance, which has resulted in interest for adapting CNNs that have already been trained on a large amount of training data to other image datasets. When adapting pretrained CNNs to new datasets, we can either use it directly as a feature extractor, a.k.a use the off-the-shelf features,~\cite{donahue2014decaf,razavian2014cnnfeatures}, or fine-tune it~\cite{Girshick2014rich-feature-hierarchies,oquab2014learning-and-transferring,pan2010transferlearning,yosinski2014transferable,Zhang2014PartbasedRCNN}. Using off-the-shelf features, we need to specify which feature representation we should extract from the network and use these for training a new classifier. Fine-tuning a CNN involves adjusting the pretrained model parameters, such that the network can e.g. classify images from a dataset different from what the CNN was trained on before. We can either choose to fine-tune the whole network or select some layer parameters to adjust while keeping the others fixed. One important factor on deciding which approach to choose is the size of the new dataset and how similar the new dataset is to the dataset which the CNN was previously trained on. A rule of thumb here is that the closer the features are to the classification layer, the features become more specific to the training data and task \cite{yosinski2014transferable}. 

Using off-the-shelf CNN features and fine-tuned CNNs have been successfully applied in \cite{donahue2014decaf,razavian2014cnnfeatures} and \cite{Girshick2014rich-feature-hierarchies, oquab2014learning-and-transferring, Zhang2014PartbasedRCNN} respectively.
In \cite{donahue2014decaf,razavian2014cnnfeatures}, it is shown that the pretrained features have sufficient representational power to generalize well to other visual recognition tasks with simple linear classifiers, such as Support Vector Machines (SVMs), without fine-tuning the parameters of the CNN to the new task. In \cite{Girshick2014rich-feature-hierarchies, Zhang2014PartbasedRCNN}, all CNN parameters are fine-tuned, whereas in \cite{oquab2014learning-and-transferring} the pretrained CNN layer parameters are kept fixed and only an adaptation layer of two fully connected layers are trained on the new task. The results from these works motivate why we should evaluate our dataset on fine-tuned CNNs or linear classifiers trained on off-the-shelf feature representations instead of training an image recognition model from scratch. 

\paragraph*{Variational Autoencoders with only natural images.}
Deep generative models, 
e.g. the variational autoencoder (VAE)~\cite{kingma2014autoencoding,Rezende2014StochasticBA,zhang2017advances}, have become widely used in the machine learning community thanks to their generative nature. We thus use VAEs for representation learning as the second benchmarking method. For efficiency, we use low-level pretrained features from a CNN as inputs to the VAE.

The latent representations from VAEs are encodings of the underlying factors for how the data are generated. VAEs belongs to the family of latent variable models, which commonly has the form $p_{\boldsymbol{\theta}}(\mathbf{x},\mathbf{z}) = p(\mathbf{z}) p_{\boldsymbol{\theta}}(\mathbf{x}|\mathbf{z})$, where $p(\mathbf{z})$ is a prior distribution over the latent variables $\mathbf{z}$ and $p_{\boldsymbol{\theta}}(\mathbf{x}|\mathbf{z})$ is the likelihood over the data $\mathbf{x}$ given $\mathbf{z}$. The prior distribution is often assumed to be Gaussian,
$p(\mathbf{z}) = \mathcal{N}(\mathbf{z}\,|\, \boldsymbol{0}, \mathbf{I})$,  
whereas the likelihood distribution depends on the values of $\mathbf{x}$.
The likelihood $p_{\boldsymbol{\theta}}(\mathbf{x}|\mathbf{z})$ is referred to as a decoder represented as a neural network parameterized by $\boldsymbol{\theta}$. An encoder network $q_{\boldsymbol{\phi}}(\mathbf{z}|\mathbf{x})$ parameterized by $\boldsymbol{\phi}$ is introduced as an approximation of the true posterior $p_{\boldsymbol{\theta}}(\mathbf{z}|\mathbf{x})$, which is intractable since it requires computing the integral $p_{\boldsymbol{\theta}}(\mathbf{x}) = \int p_{\boldsymbol{\theta}}(\mathbf{x}, \mathbf{z}) \, d\mathbf{z}$. 
When the prior distribution is a Gaussian, the approximate posterior is also modeled as a Gaussian, $q_{\boldsymbol{\phi}}(\mathbf{z}|\mathbf{x}) = \mathcal{N}(\mathbf{z} \,|\,\boldsymbol{\mu}(\mathbf{x}), \boldsymbol{\sigma}^2(\mathbf{x}) \odot \mathbf{I})$, with some mean $\boldsymbol{\mu}(\mathbf{x})$ and variance $\boldsymbol{\sigma}^2(\mathbf{x})$ computed by the encoder network. The goal is to maximize the marginal log-likelihood by defining a lower bound using $q_{\boldsymbol{\phi}}(\mathbf{z}|\mathbf{x})$:
\begin{align}
\begin{split}\label{eq:vae-loss}
\log p_{\boldsymbol{\theta}}(\mathbf{x}) \geq \mathcal{L}(\boldsymbol{\theta}, \boldsymbol{\phi}; \mathbf{x}) = & \mathbb{E}_{q_{\boldsymbol{\phi}}(\mathbf{z}|\mathbf{x})}\left[\, \log p_{\boldsymbol{\theta}}(\mathbf{x} | \mathbf{z}) \,\right] \\ & -D_{KL}(q_{\boldsymbol{\phi}}(\mathbf{z}|\mathbf{x})\,||\,p(\mathbf{z})) .
\end{split}
\end{align}
The last term is the Kullback-Leibler (KL) divergence of the approximate posterior from the true posterior. The lower bound $\mathcal{L}$ is called the evidence lower bound (ELBO) and can be optimized with stochastic gradient descent via backpropagation \cite{doersch2016tutorialvae,kingma2014autoencoding}. 
VAE is a probabilistic framework. Many extensions such as utilizing structured priors\cite{butepage2018Inform} or using continual learning \cite{nguyen2018variational} have been explored.
In the following method, we describe how to make use of the iconic images while retaining the unsupervised learning setting in VAEs.

\paragraph*{Utilizing iconic images with multi-view VAEs.}
Utilizing extra information has shown to be useful in many applications with various model designs~\cite{butepage2018Inform,vedantam2018generative,vinyals2015show,wang2016vcca,zhang2016inter}. For computer vision tasks, natural language is the most commonly used modality to aid the visual representation learning. However, the consistency of the language and visual embeddings has no guarantee. As an example with our dataset, the product description of a Royal Gala apple explains the appearance of a red apple. But if the description is represented with word embeddings, e.g. word2vec \cite{mikolov2013distributedrepresentations}, the word 'royal' will probably be more similar to the words 'king' and 'queen' than 'apple'. Therefore, if available, additional visual information about objects might be more beneficial for learning meaningful representations instead of text. In this work, with our collected dataset, we propose to utilize the iconic images for the representation learning of natural images using a multi-view VAE. Since the natural images can include background noise and grocery items different from the targeted one, the role of the iconic image will be to guide the model to which features that are of interest in the natural image.

The VAE can be extended to modeling multiple views of data, where a latent variable $\mathbf{z}$ is assumed to have generated the views \cite{vedantam2018generative,wang2016vcca}. Considering two views $\mathbf{x}$ and $\mathbf{y}$, the joint distribution over the paired random variables ($\mathbf{x}$, $\mathbf{y}$) and latent variable $\mathbf{z}$ can be written as $p_{\boldsymbol{\theta}}(\mathbf{x}, \mathbf{y}, \mathbf{z}) = p(\mathbf{z})p_{\boldsymbol{\theta^{(1)}}}(\mathbf{x}\,|\,\mathbf{z})p_{\boldsymbol{\theta^{(2)}}}(\mathbf{y}\,|\,\mathbf{z})$, where both $p_{\boldsymbol{\theta^{(1)}}}(\mathbf{x}\,|\,\mathbf{z})$ and $p_{\boldsymbol{\theta^{(2)}}}(\mathbf{y}\,|\,\mathbf{z})$ are represented as neural networks with parameters $\boldsymbol{\theta^{(1)}}$ and $\boldsymbol{\theta^{(2)}}$. Assuming that the latent variable $\mathbf{z}$ can reconstruct both $\mathbf{x}$ and $\mathbf{y}$ when only $\mathbf{x}$ is encoded into $\mathbf{z}$ by the encoder $q_{\boldsymbol{\phi}}(\mathbf{z}|\mathbf{x})$, then the ELBO is written as 
\begin{align}
\begin{split}\label{eq:vcca-loss}
\log p_{\boldsymbol{\theta}}(\mathbf{x}, \mathbf{y}) \geq & \, \mathcal{L}(\boldsymbol{\theta}, \boldsymbol{\phi}; \mathbf{x}, \mathbf{y}) \\ 
= & \,  \mathbb{E}_{q_{\boldsymbol{\phi}}(\mathbf{z}|\mathbf{x})}\left[\, \log p_{\boldsymbol{\theta^{(1)}}}(\mathbf{x} | \mathbf{z}) + \log p_{\boldsymbol{\theta^{(2)}}}(\mathbf{y} | \mathbf{z}) \,\right] \\ 
& -D_{KL}(q_{\boldsymbol{\phi}}(\mathbf{z}|\mathbf{x})\,||\,p(\mathbf{z})) .
\end{split}
\end{align}  
This model is referred to as variational autoencoder canonical correlation analysis (VAE-CCA) and was introduced in \cite{wang2016vcca}. The main motivation for using VAE-CCA is that the latent representations need to contain information about reconstructing both natural and iconic images.
The main motivation for using VAE-CCA is that 
the latent representation needs to preserve information about how both the natural and iconic images are reconstructed. This also allows us to produce iconic images from new natural images to enhance the interpretability of the latent representation of VAE-CCA (see Section \ref{sec:experimental-results}) \cite{vedantam2018generative}.

\begin{figure*}[t]
    \centering
    \subfigure[CNN]{\label{subfig:cnn}\includegraphics[width=0.35\textwidth]{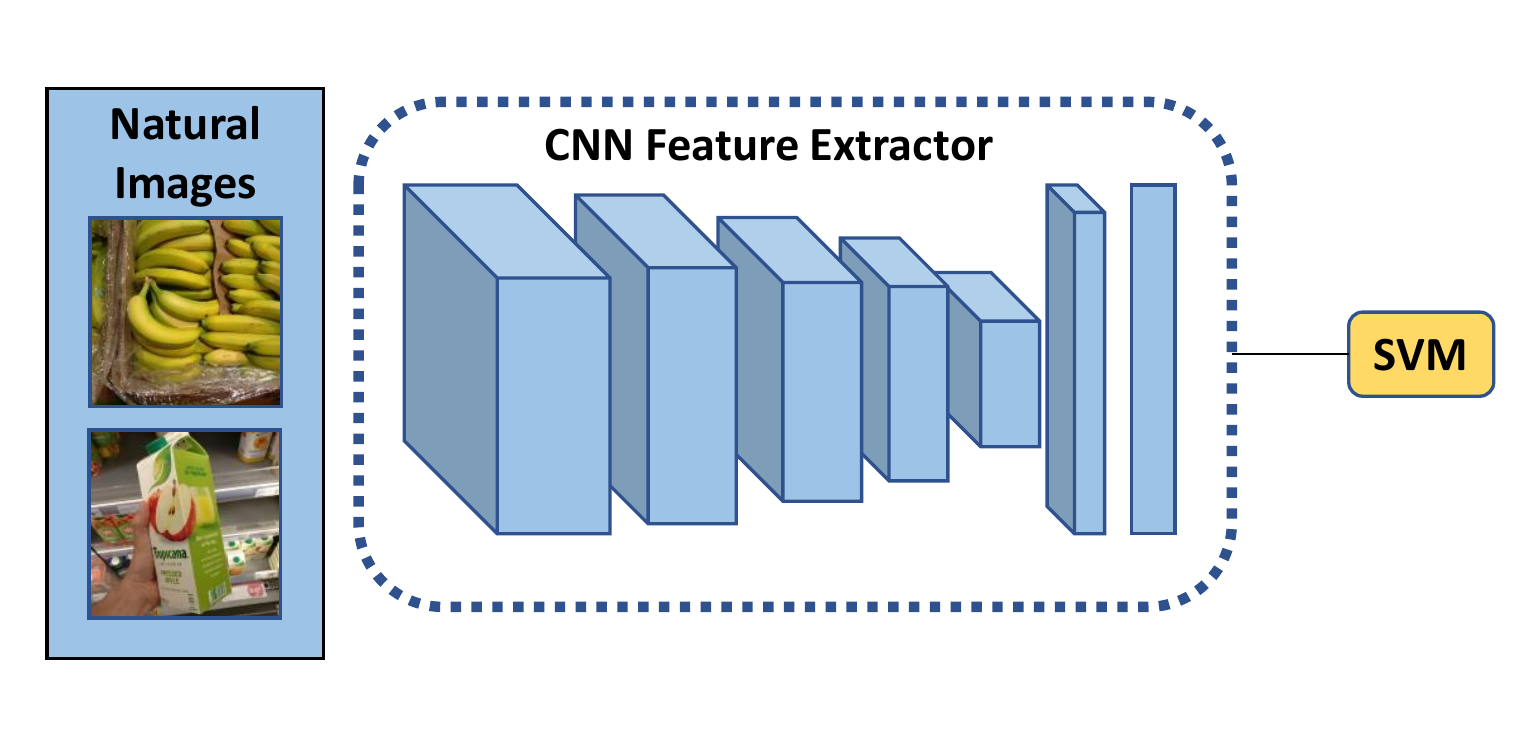}} 
    \subfigure[VAE]{\label{subfig:cnn+vae}\includegraphics[width=0.6\textwidth]{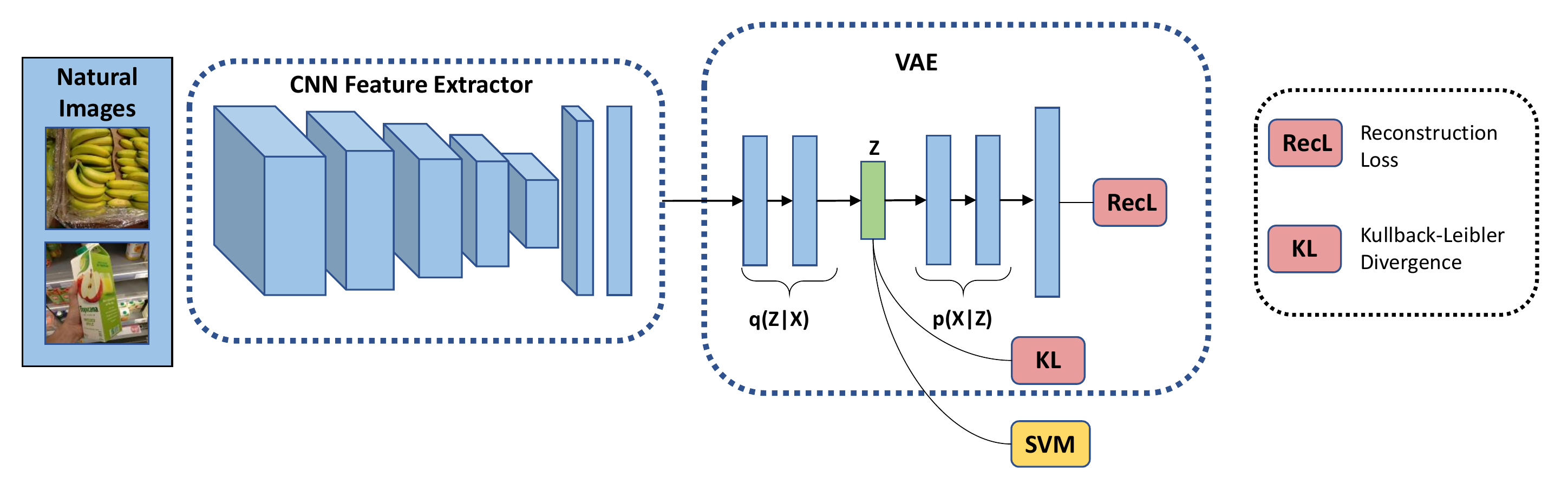}}
    \subfigure[VAE-CCA]{\label{subfig:vae-cca}\includegraphics[width=0.7\textwidth]{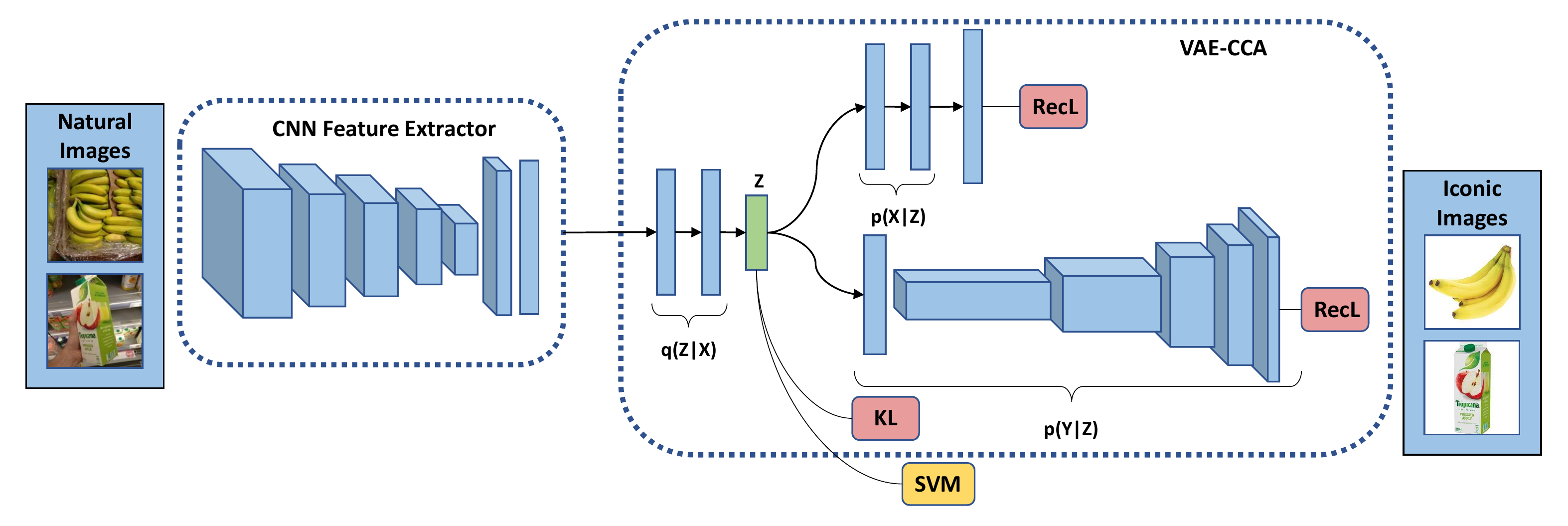}}
    \caption{The architectures for the classification methods described in Section \ref{sec:classification-methods}. In this paper, we use either a pretrained AlexNet, VGG16 or DenseNet-169 as the CNN feature extractor, but it may be replaced with any CNN architecture. Note that the pretrained CNN can be fine-tuned. The encoder and decoder of the VAE in \ref{subfig:cnn+vae} consist of two fully-connected layers. VAE-CCA in \ref{subfig:vae-cca} uses the DCGAN architecture as an iconic image decoder and the same encoder and feature vector decoder as the VAE.   }
    \label{fig:classification-methods}
\end{figure*}


\section{Experimental Results}\label{sec:experimental-results}

We apply the three different types of models described in Section \ref{sec:classification-methods} to our dataset and evaluate their performance. The natural images are propagated through a CNN pretrained on ImageNet to extract feature vectors. We experiment with both the off-the-shelf features as well as fine-tuning the CNN. When using off-the-shelf features, we simply extract feature vectors and train an SVM on those. For the fine-tuned CNN, we report both results from the softmax classifier used in the actual fine-tuning procedure and training an SVM with extracted fine-tuned feature vectors.  

These extracted feature vectors are also used for VAE and VAE-CCA which makes further compression. We perform classification for those VAE based models by training a classifier, e.g. an SVM, on the data encoded into the latent representation. We use this classification approach for both VAE and VAE-CCA. In all classification experiments, except when we fine-tune the CNN, we use a linear SVM trained with the one-vs-one approach as in \cite{razavian2014cnnfeatures}.

We experiment with three different pretrained CNN architectures, namely AlexNet \cite{krizhevsky2012imagenet}, VGG16 \cite{simonyan2014verydeep} and DenseNet-169 \cite{huang2017densely}. For AlexNet and VGG16, we extract feature vectors of size 4096 from the two last fully connected (FC) layers before the classification layer. The features from the $n^{\text{th}}$ hidden layer are denoted as $\text{AlexNet}_{n}$ and $\text{VGG16}_{n}$. As an example, the last hidden FC layer in AlexNet is denoted as $\text{AlexNet}_{7}$, the input of which is output from $\text{AlexNet}_{6}$. For DenseNet-169, we extract the features of size 1664 from the average pooling layer before its classification layer.

\renewcommand{\arraystretch}{1.2}
\begin{table*}[t]
    \centering
    \caption{ Fine-grained classification (81 classes) accuracies with the methods described in Section \ref{subsec:experimental-setups}. Each row displays from which network architecture and layer that we extracted the feature vectors of the natural images. The columns show the result from the classifiers that we used (see Section \ref{subsec:experimental-setups}). }
    \vspace{2mm}
    \begin{tabular}{c | c c | c c | c c}
        \hline
        \Xhline{3\arrayrulewidth}
        & SVM & SVM-ft & VAE+SVM & VAE+SVM-ft & VAE-CCA+SVM & VAE-CCA+SVM-ft  \\
        \Xhline{3\arrayrulewidth}
        \rowcolor{Gray}
         $\text{AlexNet}_{6}$  &  69.2 & 72.6 & 65.6 & 70.7 & 67.8	& 71.5 \\
         $\text{AlexNet}_{7}$ &  65.0 & 70.7 & 63.0 & 68.7 & 65.0 & 70.9 \\
        \rowcolor{Gray}
         $\text{VGG16}_{6}$ &  62.1	& 73.3 & 57.5 & 71.9 & 60.7 & 73.0 \\
         $\text{VGG16}_{7}$ & 57.3	& 71.7 & 56.8 & 67.8 & 56.8 & 71.3 \\
        \rowcolor{Gray}
         $\text{DenseNet-169}$ & 72.5 & 85.0 & 65.4 & 79.1 & 72.6 & 80.4 \\
        \Xhline{3\arrayrulewidth}
    \end{tabular}
    \label{tab:results-fine-grained}
\end{table*}

\renewcommand{\arraystretch}{1.2}
\begin{table}[t]
    \centering
    \caption{Coarse-grained classification (46 classes) accuracies with an SVM for the methods described in Section \ref{subsec:experimental-setups} that uses off-the-shelf feature representations. Each row displays from network architecture and layer that we extracted the feature vectors of the natural images and the columns show the result for the classification methods. }
    \vspace{2mm}
    \scalebox{0.95}{
    \begin{tabular}{c | ccc}
        \Xhline{3\arrayrulewidth}
         & SVM & VAE+SVM & VAE-CCA+SVM  \\
        \Xhline{3\arrayrulewidth}
        \rowcolor{Gray}
         $\text{AlexNet}_{6}$ & 78.0 & 74.2 & 76.4  \\ 
         $\text{AlexNet}_{7}$ & 75.4 & 73.2 & 74.4  \\ 
        \rowcolor{Gray}
         $\text{VGG16}_{6}$ &  76.6 & 74.2 & 74.9  \\ 
         $\text{VGG16}_{7}$ & 72.8 & 71.7 & 72.3  \\
         \rowcolor{Gray}
         $\text{DenseNet-169}$ & 85.2 & 79.5 & 82.0 \\
        \Xhline{3\arrayrulewidth}
    \end{tabular}
    }
    \label{tab:results-coarse-grained}
\end{table}

\renewcommand{\arraystretch}{1.25}
\begin{table}[t]
\centering
\caption{Fine-grained classification accuracies from fine-tuned CNNs pretrained on ImageNet, where the column shows which architecture that has been fine-tuned. A standard softmax layer is used as the last classification layer.
}
\vspace{2mm}
\scalebox{0.95}{
\begin{tabular}{cccc}
    \hline
    \Xhline{3\arrayrulewidth}
    & AlexNet & VGG16 & DenseNet-169 \\
    \Xhline{3\arrayrulewidth}
    \rowcolor{Gray}
    Fine-tune &  69.3 & 73.8 & 84.0  \\
    \Xhline{3\arrayrulewidth}
\end{tabular}
}
\label{tab:results-finetuned-cnn}
\end{table}

\subsection{Experimental Setups}\label{subsec:experimental-setups}

The following setups were used in the experiments:

\paragraph*{Setup 1.} Train an SVM on extracted off-the-shelf features from a pretrained CNN, which is denoted as SVM in the results. We also fine-tune the CNN by replacing the final layer with a new softmax layer and denote these results as Fine-tune. We denote training an SVM on extracted finetuned feature vectors as SVM-ft.

\paragraph*{Setup 2.} Extract feature vectors with a pretrained CNN of the natural images and learn a latent representation $\mathbf{z}$ with a VAE. Then the data is encoded into the latent space and we train an SVM with these latent representations, which used for classification. We denote the results as VAE+SVM when using off-the-shelf feature vectors, whereas using the fine-tuned feature vectors are denoted as VAE+SVM-ft. In all experiments with the VAE, we used the architecture from \cite{sohn2015conditionalvae}, i.e. the latent layer having 200 hidden units and both encoder and decoder consisting of two FC layers with 1,000 hidden units each.

\paragraph*{Setup 3.} Each natural image is paired with its corresponding iconic image. We train VAE-CCA similarly as the VAE, but instead, we learn a joint latent representation that is used to reconstruct the extracted feature vectors $\mathbf{x}$ and the iconic images $\mathbf{y}$. The classification is performed with the same steps as in Setup 2 and denotes the results similarly with VAE-CCA+SVM and VAE-CCA+SVM-ft. Our VAE-CCA model takes the feature vectors $\mathbf{x}$ as input and encodes them into a latent layer with 200 hidden units. The encoder and the feature vector decoder uses the same architecture, i.e. two FC layers with 512 hidden units, whereas the iconic image decoder uses the DCGAN \cite{radford2015unsupervised} architecture.

Figure \ref{fig:classification-methods} displays the three experimental setups described above. We report both fine-grained and coarse-grained classification results with an SVM in Table \ref{tab:results-fine-grained} and \ref{tab:results-coarse-grained} respectively. In Table \ref{tab:results-finetuned-cnn}, we report the fine-grained classification results from fine-tuned CNNs.

When fine-tuning the CNNs,
we replace the final layer with a softmax layer applicable to our dataset with randomly initialized weights drawn from a Gaussian with zero mean and standard deviation $0.01$~\cite{Zhang2014PartbasedRCNN}. 
For AlexNet and VGG16, we fine-tune the networks for 30 epochs with two different learning rates, 0.01 for the new classification layer and 0.001 for the pretrained layers. Both learning rates are reduced by half after every fifth epoch. The DenseNet-169 is fine-tuned for 30 epochs with momentum of $0.9$ and an initial learning rate of 0.001, which decays with $10^{-6}$ after each epoch. We report the classification results from the softmax activation after the fine-tuned classification layer. 
We also report classification results from an SVM trained with feature representations from a fine-tuned CNN, which are extracted from FC6 and FC7 of the AlexNet and VGG16 and from the last average pooling layer in DenseNet-169.

The VAE and VAE-CCA models are trained for 50 epochs with Adam \cite{kingma2015adam} for optimizing the ELBOs in Equation \ref{eq:vae-loss} and \ref{eq:vcca-loss} respectively. We use a constant learning rate of 0.0001 and set the minibatch size to 64. The extracted feature vectors are rescaled with standardization before training the VAE and VAE-CCA models to stabilize the learning.

\subsection{Results}
\begin{figure*}[t]
\centering
\subfigure[Royal Gala]{\label{subfig:royal-gala-natural}\includegraphics[width=0.33\columnwidth]{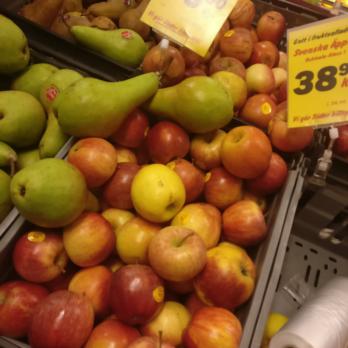}}~
\subfigure[Decoded Royal Gala]{\label{subfig:royal-gala-decoded}\includegraphics[width=0.33\columnwidth]{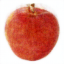}}~
\subfigure[Brown Cap]{\label{subfig:brown-cap-natural}\includegraphics[width=0.33\columnwidth]{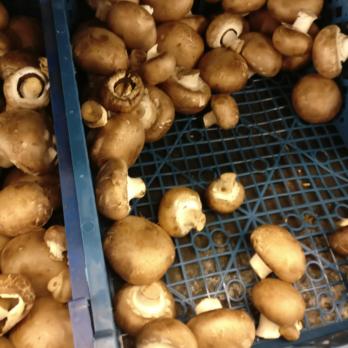}}~
\subfigure[Decoded\,Brown\,Cap]{\label{subfig:brown-cap-decoded}\includegraphics[width=0.33\columnwidth]{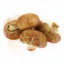}}~
\subfigure[Oatgurt]{\label{subfig:oatgurt-natural}\includegraphics[width=0.33\columnwidth]{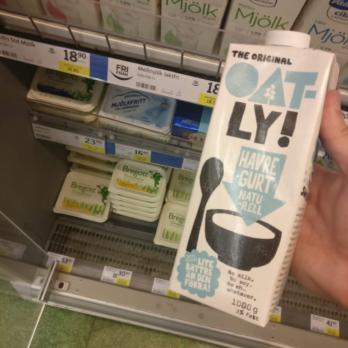}}~
\subfigure[Decoded Oatgurt]{\label{subfig:oatgurt-decoded}\includegraphics[width=0.33\columnwidth]{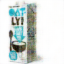}}~ \\
\subfigure[Orange]{\label{subfig:orange-natural}\includegraphics[width=0.33\columnwidth]{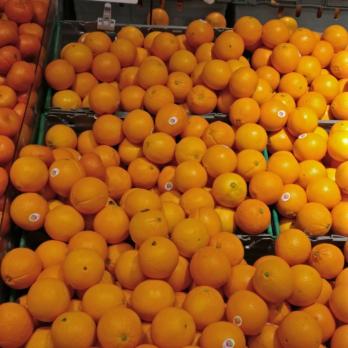}}~
\subfigure[Decoded Orange]{\label{subfig:orange-decoded}\includegraphics[width=0.33\columnwidth]{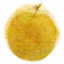}}~
\subfigure[Onion]{\label{subfig:onion-natural}\includegraphics[width=0.33\columnwidth]{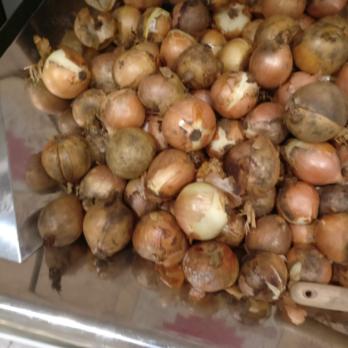}}~
\subfigure[Decoded Onion]{\label{subfig:onion-decoded}\includegraphics[width=0.33\columnwidth]{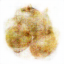}}~
\subfigure[Yogurt]{\label{subfig:yogurt-natural}\includegraphics[width=0.33\columnwidth]{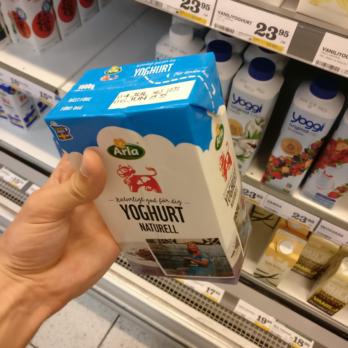}}~
\subfigure[Decoded Yogurt]{\label{subfig:yogurt-decoded}\includegraphics[width=0.33\columnwidth]{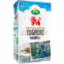}}~
\caption{ Examples of natural images in the test set that have been decoded into product iconic images by the iconic image decoder. This result is obtained with the fine-tuned DenseNet-169 features, which corresponds to VAE-CCA+SVM-ft in Table \ref{tab:results-fine-grained}. Subfigures (a), (c), (e), (g), (i) and (k) show the example input image from the test set, and Subfigures (b), (d), (f), (h), (j) and (l) show the decoded iconic image from the decoder $p_{\boldsymbol{\theta^{(2)}}}(\mathbf{y}\,|\,\mathbf{z})$ using VAE-CCA model as in Figure \ref{subfig:vae-cca}.} \label{fig:decoded-images}
\end{figure*}

The fine-grained classification results for all methods using an SVM as classifier are shown in Table \ref{tab:results-fine-grained}. We also provide coarse-grained classification results for some of the methods in Table \ref{tab:results-coarse-grained} to demonstrate the possibility of hierarchical evaluation that our labeling of the data provides (see Figure \ref{fig:examples}). The accuracies in the coarse-grained classification are naturally higher than the accuracies in the corresponding columns in Table \ref{tab:results-fine-grained}. Table \ref{tab:results-finetuned-cnn} shows fine-grained classification accuracies from a softmax classifier in the fine-tuned CNNs. We note that fine-tuning the networks gives consistently better results than training an SVM on off-the-shelf features (see Table \ref{tab:results-fine-grained}).

Fine-tuning the entire network results improves the classification performance consistently for each method in Table \ref{tab:results-fine-grained}. The performance is clearly enhanced for features extracted from fine-tuned VGG16 and DenseNet-169, which improves the classification accuracy by 10\% in most cases for SVM-ft, VAE+SVM-ft, and VAE-CCA+SVM-ft. For AlexNet and VGG16, we see that the performance drops when extracting the features from layer FC7 instead of FC6. The reason might be that the off-the-shelf features in FC7 are more difficult to transfer to other datasets since the weights are biased towards classifying objects in the ImageNet database. The performance drops also when we use fine-tuned features, which could be due to the small learning rate we use for the pretrained layers, such that the later layers are still ImageNet-specific. We might circumvent this drop by increasing the learning rate for the later pretrained layers and keeping the learning rate for earlier layers small.
 
The VAE-CCA model achieves mostly higher classification accuracies than the VAE model in both Table \ref{tab:results-fine-grained} and \ref{tab:results-coarse-grained}. This indicates that the latent representation separates the classes more distinctly than the VAE by jointly learning to reconstruct the extracted feature vectors and iconic images. However, further compressing the feature vectors with VAE and VAE-CCA will lower the classification accuracy compared to applying the feature vectors to a classifier directly. Since both VAE and VAE-CCA compresses the feature vectors into the latent representation, there is a risk of losing information about the natural images. We might receive better performance by increasing the dimension of the latent representation at the expense of speed in both training and classification.

In Figure \ref{fig:decoded-images}, we show results from the iconic image decoder $p_{\boldsymbol{\theta^{(2)}}}(\mathbf{y}\,|\,\mathbf{z})$ when translating natural images from the test set into iconic images with VAE-CCA and a fine-tuned DenseNet-169 as feature extractor. Such visualization can demonstrate the quality of the representation using the model, as well as enhancing the interpretability of the method. Using VAE-CCA in the proposed manner, we see that with challenging natural images, the model is still able to learn an effective representation which can be decoded to the correct iconic image. For example, some pears have been misplaced in the bin for Royal Gala apples in Figure \ref{subfig:royal-gala-natural}, but still the image decoder manages to decode a blurry red apple seen in Figure \ref{subfig:royal-gala-decoded}. In Figure \ref{subfig:orange-decoded}, a mix of an orange and an apple are decoded from a bin of oranges in Figure \ref{subfig:orange-natural}, which indicates these fruits are encoded close to each other in the learned latent space. Even if Figure \ref{subfig:oatgurt-natural} includes much of the background, the iconic image decoder is still able to reconstruct the iconic images accurately in Figure \ref{subfig:oatgurt-decoded}, which illustrates that the latent representation is able to explain away irrelevant information in the natural image and preserved the features of the oatgurt package. Thus, using VAE-CCA with iconic images as the second view not only advances the classification accuracy but also provides us with the means to understand the model.

\section{Conclusions}\label{sec:conclusions}

This paper presents a dataset of images of various raw and packaged grocery items, such as fruits, vegetables, and dairy and juice products. We have used a structured labeling of the items, such that grocery items can be grouped into more general (coarse-grained) classes and also divided into fine-grained classes. For each class, we have a clean iconic image and a text description of the item, which can be used for adding visual and semantic information about the items in the modeling. The intended use of this dataset is to train and benchmark assistive systems for visually impaired people when they shop in a grocery store. Such a system would complement existing visual assistive technology, which is confined to grocery items with barcodes. We also present preliminary benchmark results for the dataset on the task of image classification. 

We make the dataset publicly available for research purposes at \url{https://github.com/marcusklasson/GroceryStoreDataset}. Additionally, we will both continue collecting natural images, as well as ask for public contributions of natural images in shopping scenarios to enlarge our dataset. 

For future research, we will advance our model design to utilize the structured nature of our labels. Additionally, we will design a model that use the product description of the objects in addition to the iconic images. One immediate next step is to extend the current VAE-CCA model to three views, where the third view is the description of the product.

{\small
\bibliographystyle{ieee}
\bibliography{ref}
}

\end{document}